\documentclass{article}

\usepackage[numbers]{natbib}
\usepackage[preprint]{neurips_2023}

\usepackage[utf8]{inputenc} % allow utf-8 input
\usepackage[T1]{fontenc}    % use 8-bit T1 fonts
\usepackage{hyperref}       % hyperlinks
\usepackage{url}            % simple URL typesetting
\usepackage{booktabs}       % professional-quality tables
\usepackage{amsfonts}       % blackboard math symbols
\usepackage{nicefrac}       % compact symbols for 1/2, etc.
\usepackage{microtype}      % microtypography
\usepackage{xcolor}         % colors
\usepackage{graphicx}
\usepackage{lipsum}
\usepackage{amsmath, amsthm, amssymb}
\usepackage{floatrow}
\usepackage{pifont}
\newcommand{\cmark}{\ding{51}}
\newcommand{\xmark}{\ding{55}}
\usepackage{caption}
\usepackage{array}
\usepackage{makecell}
\usepackage{xfrac}
\usepackage{mathtools}
\usepackage{multirow}

\definecolor{darkgreen}{RGB}{50, 150, 50}
\title{HomE: Homography-Equivariant \\Video Representation Learning} 

\author{Anirudh Sriram$^1$, Adrien Gaidon$^2$, Jiajun Wu$^1$, Juan Carlos Niebles$^1$, Li Fei-Fei$^1$, Ehsan Adeli$^{1,3}$ \\
\footnotesize $^1$Department of Computer Science, Stanford University, Stanford, CA 94305, USA \\ 
$^2$Toyota Research Institute, Los Altos, CA 94022, USA \\ 
\footnotesize $^3$Department of Psychiatry and Behavioral Sciences, Stanford University, Stanford, CA 94305, USA \\ 
\footnotesize \{sanirudh,\, jiajunwu,\, jniebles,\, feifeili,\, eadeli\}@cs.stanford.edu  \& \footnotesize adrien.gaidon@tri.global}

\begin{document}
 
\maketitle

\begin{abstract}
Recent advances in self-supervised representation learning have enabled more efficient and robust model performance without relying on extensive labeled data. However, most works are still focused on images, with few working on videos and even fewer on multi-view videos, 
where more powerful inductive biases can be leveraged for self-supervision.
% where more powerful information can be leveraged through self-supervision.
In this work, we propose a novel method for representation learning of multi-view videos, where we explicitly model the representation space to maintain Homography Equivariance (HomE). 
% Specifically, we learn a representation of multi-view videos that is equally variant with respect to the homography operation. 
Our method learns an implicit mapping between different views, culminating in a representation space that maintains the homography relationship between neighboring views.  We evaluate our HomE representation via action recognition and  pedestrian intent prediction as downstream tasks. On action classification, our method obtains 96.4\% 3-fold accuracy on the UCF101 dataset, better than most state-of-the-art self-supervised learning methods. Similarly, on the STIP dataset, we outperform the state-of-the-art by 6\% for pedestrian intent prediction one second into the future while also obtaining an accuracy of 91.2\% for pedestrian action (cross vs. not-cross) classification. Code is available at {\small\url{https://github.com/anirudhs123/HomE}}.

% \note{
%   The abstract paragraph should be indented \nicefrac{1}{2}~inch (3~picas) on
%   both the left- and right-hand margins. Use 10~point type, with a vertical
%   spacing (leading) of 11~points.  The word \textbf{Abstract} must be centered,
%   bold, and in point size 12. Two line spaces precede the abstract. The abstract
%   must be limited to one paragraph.}
\end{abstract}

\section{Introduction}
\label{Introduction}
Over the past few years, there has been tremendous progress in self-supervised methods revolutionizing the field of representation learning \cite{DBLP:conf/icml/ChenK0H20, DBLP:journals/corr/abs-2010-09709, Tao2020SelfSupervisedVR}. 
Obtaining high-quality manual annotations are costly and time-consuming, hence supervised learning has become less appealing, especially when considering tasks that involve complex data like videos. Video is a natural domain for representation learning methods, since data is rich and abundant \cite{DBLP:journals/corr/BilenFGV16, 10.1145/1553374.1553469, NIPS2009_043c3d7e}, but hard to annotate at scale due to the additional temporal complexity. These self-supervised learning methods can successfully leverage large amounts of uncurated data to improve representations. This has led to self-supervised representations often comparable to or even outperforming \citep{DBLP:journals/corr/abs-1911-05722} supervised techniques on certain downstream tasks. 
Video representation learning plays a crucial role in many downstream tasks like action recognition, forecasting, segmentation, and many others~\cite{DBLP:journals/corr/FernandoBGG16,DBLP:journals/corr/Abu-El-HaijaKLN16, 6909619, 6126543}. 
%As pointed out in \cite{DBLP:journals/corr/FernandoBGG16}, videos, unlike images, are generally open-ended media, and one cannot know beforehand a frame range within which we can expect an action to occur or an object to be found. We would then need to manually annotate videos frame-by-frame or crop them to a range of frames to ensure consistency. Also, large video datasets that are used commonly like \cite{DBLP:journals/corr/Abu-El-HaijaKLN16, 6909619, 6126543}, rely on noisy, unreliable tags. As a result, one cannot truly point a finger at the architecture or the noisy labels that contribute to the observed network behavior, thus needing video based self-supervised techniques now more than ever. 

\begin{figure}[!]
         \centering
        \includegraphics[width = 0.95\columnwidth]{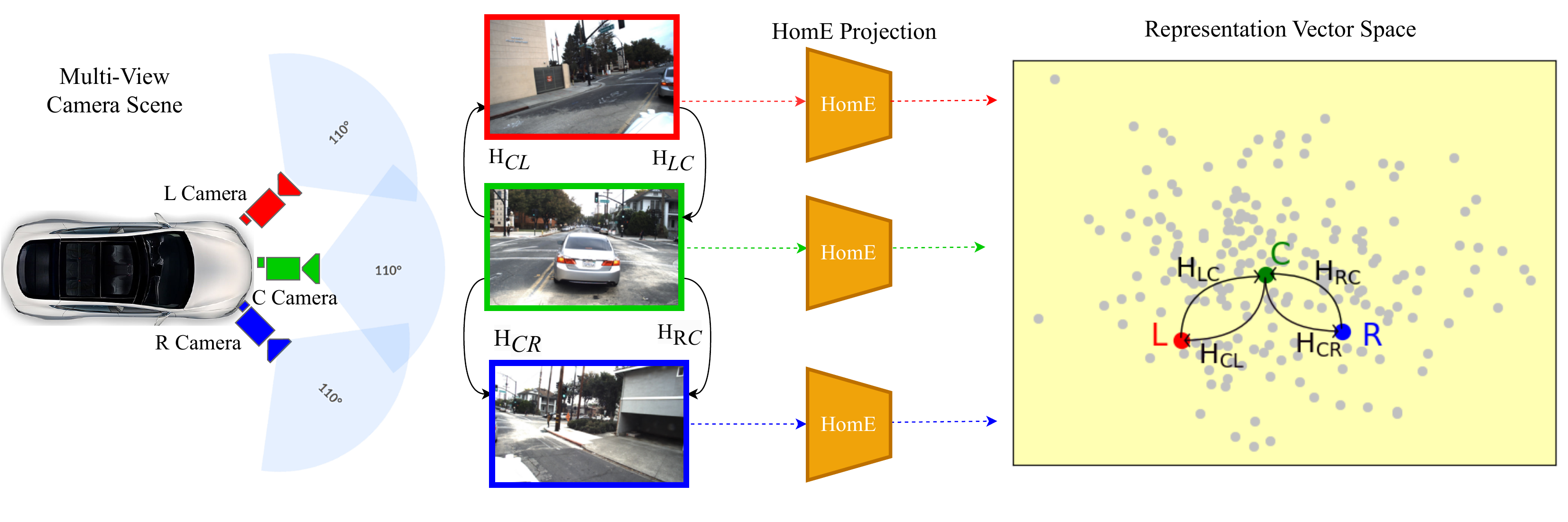} \vspace{-15pt}
        \caption{Unlike other multi-view representation learning methods, we do not align the representations of different views; instead, we explicitly model and learn  representations that are equivariant with respect to the Homography operation. L: Left, C: Center, R: Right, $\mathbf{H}$: Homography matrices between each pair of neighboring views. }
        \label{fig:HomE_Gen_Concept}
    \end{figure}

Self-supervised learning methods often use intrinsic and structural properties of the data as a supervisory signal to avoid the need for %, so these methods are unsupervised in the sense that it does not require 
human annotation. %but supervised machine learning techniques can still be used. 
Video-based self-supervised learning methods can be divided in two broad categories: \textit{contrastive learning methods} and \textit{pretext task methods}. In \textit{contrastive learning methods} \cite{DBLP:journals/corr/abs-1805-01978, DBLP:journals/corr/abs-1807-03748, DBLP:journals/corr/abs-1906-05849}, the key idea is instance discrimination. 
Generally, samples from the same video, either different modalities or different crops of the same video are treated as positive samples, while samples from different videos are considered negative. The networks learn to distinguish one instance from another. 
% Thus, the representations learned would be sufficient for downstream tasks. 
\textit{Pretext task methods} rely on designing learning objectives that encode structural assumptions about the data to constrain models to learn 
% effective and 
informative representations. For video data, some of these pretext tasks include detecting image rotation angles \cite{DBLP:journals/corr/abs-1803-07728}, 
% image in-painting \cite{DBLP:journals/corr/PathakKDDE16}, and
solving jigsaw puzzles\cite{DBLP:journals/corr/NorooziF16}, temporal tasks such as predicting video clip orders \cite{DBLP:journals/corr/abs-1708-01246, 8953292} or frame orders \cite{DBLP:journals/corr/MisraZH16}, video playback speed \cite{DBLP:journals/corr/abs-2003-02692, DBLP:journals/corr/abs-2008-05861}, temporal transformations \cite{DBLP:journals/corr/abs-2007-10730} and video reconstruction \cite{tong2022videomae}.

Multi-view videos, such as those used in autonomous driving applications (see Fig.~\ref{fig:HomE_Gen_Concept}), can provide some additional sources of supervision for contrastive learning \cite{DBLP:journals/corr/abs-1906-05849, 8014803}. These works align the representations from different views, minimizing the distance between the representations of each pair of views in the learned space. However, this objective might be too strict and noisy in practice, as semantics might significantly vary across views, especially if they have limited overlap (e.g., one center camera observing the front of the vehicle and the right camera mainly having the sidewalk in its view).

Instead, we propose a more principled inductive bias rooted in multi-view geometry: \emph{Homography-Equivariant Video Representation Learning (HomE)}.
As shown in Fig.~\ref{fig:HomE_Gen_Concept}, each training example comprises a pair of input frames from a multi-view camera setting. We employ the homography transformation between the given pairs of views and pass the video from each view through our HomE encoder. It results in a vector space in which the representations are defined as vector neurons \cite{DBLP:journals/corr/abs-2104-12229}, and hence the homography transformation can similarly be applied to them. We train HomE such that the representations maintain the same homography correspondence as the one between the images themselves.
% With HomE, we could obtain performances better than or comparable to several SOTA models on the selected downstream tasks of action classification and pedestrian intent prediction. 

As HomE operates on the frame level, we validate our hypothesis about the efficiency of learning an equivariant representation first on image classification with a synthetic dataset created out of CIFAR-10 \cite{cifar10}.  We then evaluate our method on the UCF-101 \cite{DBLP:journals/corr/abs-1212-0402} dataset by synthetically generating multiple views on the action classification task. Finally, since the multi-view setup is common in autonomous driving applications, we test HomE for the downstream task of pedestrian intent prediction task by training our model on the Stanford-TRI Intent Prediction (STIP) dataset \cite{DBLP:journals/corr/abs-2002-08945}. Our representation outperforms the state-of-the-art (SOTA) models on these different benchmarks, proving the effectiveness and the generality of our proposed representation learning technique. 

Our contributions in this paper are threefold. First, we propose a simple and efficient representation learning technique that learns a vector space preserving the spatial structure between input views and learned representations.
% using a novel pretext task. 
Unlike other multi-view representation learning methods, we do not align the representations of different views, we rather learn representations that are related to one another through their Homography matrices. Second, we also develop a neural network model consisting of an encoder, vector neuron layers \cite{DBLP:journals/corr/abs-2104-12229}, and a decoder network to assist the learning. Third, we improve the performance on pedestrian intent prediction on the STIP dataset. We further validate our algorithm for action classification on the UCF-101 dataset and image classification on the synthetic CIFAR-10 dataset, achieving SOTA.

\section{Related work} 
\textbf{\textit{Self-supervised learning with Images:}} Most self-supervised methods on images learn a representation by defining a pretext task, which is a supervised task designed to use the structure of the input data to learn useful representations. Spatial pretext tasks include predicting the position of the patches relative to one another \cite{DBLP:journals/corr/DoerschGE15}, image rotations \cite{DBLP:journals/corr/abs-1803-07728}, solving jigsaw puzzles \cite{DBLP:journals/corr/NorooziF16}, image in-painting \cite{DBLP:journals/corr/PathakKDDE16}. 
% These pretext tasks guide the model to learn representations useful for their corresponding downstream tasks.

Contrastive learning methods \cite{hjelm2019learning,DBLP:journals/corr/abs-1805-01978, DBLP:journals/corr/abs-1906-05849, DBLP:journals/corr/abs-2003-04297, DBLP:journals/corr/abs-1911-05722, 
DBLP:conf/icml/ChenK0H20,
DBLP:journals/corr/abs-1807-03748} also originated with images, where the network tries to learn by pulling representations of similar images (positive pairs) closer and pushing representations of different images (negative pairs) further apart \cite{DBLP:journals/corr/abs-1906-00910, DBLP:journals/corr/abs-1807-03748}. 
% In SimCLR \cite{10.5555/3524938.3525087}, augmented version of the same image were considered positive pairs while augmented versions of other images were considered negative pairs. Contrastive predictive coding (CPC) \cite{DBLP:journals/corr/abs-1807-03748} attempted to learn the future from the past by using sequential data. 
The main drawbacks of contrastive approaches are that they require a careful choice of positive and negative pairs \cite{DBLP:journals/corr/abs-2005-10243} and that they often rely on a large number of such negative pairs, inducing a high computational cost \cite{DBLP:conf/icml/ChenK0H20}. \cite{DBLP:journals/corr/abs-2006-07733} came up with a way to tackle one of the limitations, where the network is trained without negative samples. Alternatives to the contrastive approach, such as clustering \cite{DBLP:journals/corr/abs-1911-12667, DBLP:journals/corr/abs-1911-05371, DBLP:journals/corr/BautistaSSO16, DBLP:journals/corr/abs-1807-05520, DBLP:journals/corr/abs-2006-09882, DBLP:journals/corr/abs-1904-11567} and regression \cite{DBLP:journals/corr/abs-2002-12247, richemond2020byol}, avoid the need and cost of multiple negatives.

% Clustering-based approaches like \cite{DBLP:journals/corr/abs-1911-12667, DBLP:journals/corr/abs-1911-05371, DBLP:journals/corr/BautistaSSO16, DBLP:journals/corr/abs-1807-05520, DBLP:journals/corr/abs-2006-09882, DBLP:journals/corr/abs-1904-11567} generally try to learn unsupervised cluster assignments as targets along with the neural network parameters trained for the supervised task or they try to perform clustering using current representations (online or offline). Regression-based methods try to directly regress a representation extracted from a different view of the image \cite{DBLP:journals/corr/abs-2002-12247, richemond2020byol}.

\textbf{\textit{Self-supervised learning with Videos:}} In the video domain, spatial tasks used on images for self-supervision have still been very effective, such as completing space-time cubic puzzles \cite{DBLP:journals/corr/abs-1811-09795} and predicting video rotation angles \cite{DBLP:journals/corr/abs-1811-11387}. Compared to image data, videos have an additional temporal component that is put to use in many recent pretext task choices \cite{DBLP:journals/corr/abs-1708-01246, DBLP:journals/corr/MisraZH16, DBLP:journals/corr/FernandoBGG16}. These tasks included learning to predict the correct order of shuffled frames,  
% learning to distinguish whether the input frames were in the correct temporal order or 
train a network to identify unrelated or odd video clips.
% and in \cite{8578938}, the network predicts the arrow of time. 
% These tasks learn to focus on learning representations in a more abstract space.
In \cite{DBLP:journals/corr/PatrauceanHC15,
DBLP:journals/corr/abs-1806-09594,
DBLP:journals/corr/MathieuCL15,
DBLP:journals/corr/SrivastavaMS15,
DBLP:journals/corr/VondrickPT16} the networks were trained to predict future frames in pixel space by minimizing mean squared error (MSE) loss or adversarial losses. The predictions of these models are usually blurred and cannot go beyond predicting short clips into the future. 
Many recent works have started to utilize the playback speed of the input video clips. \cite{DBLP:journals/corr/abs-2003-02692} trained a network to sort video clips according to the corresponding playback rates, while 
% Playback rate perception (PRP) 
\cite{DBLP:journals/corr/abs-2006-11476} roots in a sampling strategy, which produces information about playback rates for representation learning. 
% SpeedNet \cite{DBLP:journals/corr/abs-2004-06130} was trained to detect whether a video is playing at a normal or sped-up pace.

Video contrastive methods have been very successful too \cite{DBLP:journals/corr/abs-2007-13278,
DBLP:journals/corr/abs-2008-03800,
DBLP:journals/corr/abs-2010-09709}. Samples from the same video, either different modalities or 
 % different crops or 
different augmented versions are treated as positive samples, while samples from different videos are considered negative samples. In addition, several works use temporal cues to build positive pairs as well. 
% Contrastive multiview coding (CMC) \cite{DBLP:journals/corr/abs-1906-05849} used different views (e.g. different color spaces) from the same sample as positive pairs, which can be adapted to videos with ease. 
% DPC 
\cite{DBLP:journals/corr/abs-1909-04656} and 
% MemDPC
\cite{DBLP:journals/corr/abs-2008-01065} are predictive coding methods like 
% CPC 
\cite{DBLP:journals/corr/abs-1807-03748} which were proposed to handle video data. 
% Inter-intra constrastive learning (IIC) 
\cite{DBLP:journals/corr/abs-2008-02531} introduced intra-negative video samples to enhance contrastive learning. Most of these contrastive learning methods utilize a noise contrastive estimation loss \cite{pmlr-v9-gutmann10a} for robust and effective training. Yet the costs of training such systems are significant, and complex hard-negative mining strategies are needed to improve the training efficiency \cite{DBLP:journals/corr/FaghriFKF17}.

\textbf{\textit{Multi-view representation learning:}} The heart of the problem we are trying to solve is to come up with a simple and efficient representation for multi-view data. In
% Contrastive multi-view coding (CMC) 
\cite{DBLP:journals/corr/abs-1906-05849}, representation is learned by bringing views of the same scene together in embedding space, while pushing views of different scenes apart. In 
% Time-Contrastive Networks 
\cite{8014803}, they train a representation using a metric learning loss, where multiple simultaneous viewpoints of the same observation are attracted in the embedding space while being repelled from temporal neighbors which are often visually similar but functionally different. We do not intend to come up with yet another method for video representation learning, but rather build an intuitive vector space where representations of the same scene from different views are not aligned on top of one another, unlike \cite{DBLP:journals/corr/abs-1906-05849, 10031025,8014803, ZHANG2022102160}, but are related to one another through their Homography matrices. 
% The learned representations perform comparable and sometimes better than the SOTA models.
\label{related_works}

\vspace{-5pt}
\section{Homography-Equivariant (HomE) Representation Learning}
\label{methodology}
% \subsection{Motivation} 
% Given a pair of images from two different views of the same scene, the minimum information we possess is the Homography matrix between them. And the questions that we ask are: 1) Will it be effective and useful to learn representations for them that are related to one another instead of bringing them as close as possible? 2) Can the Homography matrix, the minimum information we possess be that relation?
% \subsection{} 
The goal of self-supervised video representation learning is to learn effective feature representations from videos using a backbone network. Homography equivariant representation learning (HomE) learns to come up with representations for different views that are related to one another through their Homography matrices. HomE operates on a frame level and hence can be used not just for video representation learning but can also be extended to any multi-view setting.

\begin{figure}[!]
         \centering
        \includegraphics[width = \columnwidth]{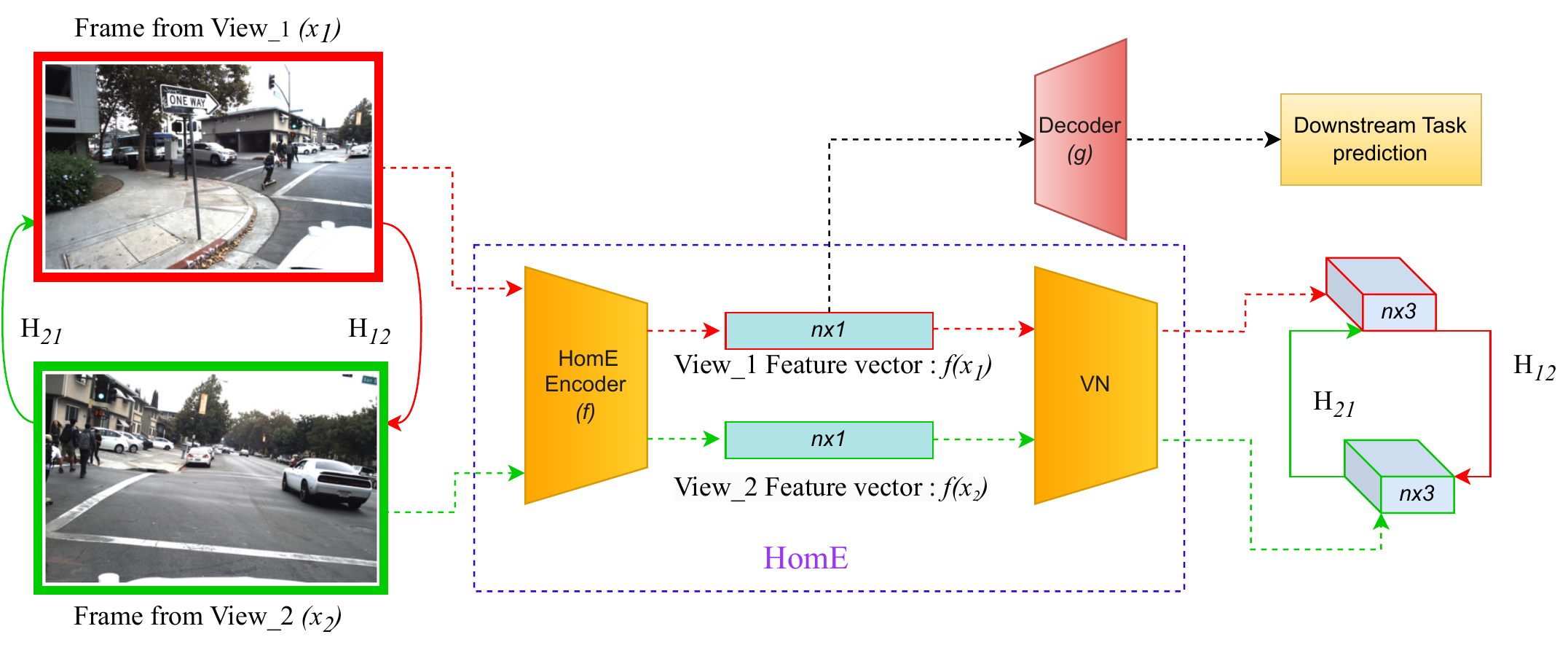} \vspace{-10pt}
        \caption{Overview of the model: The model takes in frames from different views $x_1, x_2, \ldots, x_N$ 
        % (where $N$ is the number of views) 
        as input and learns a representation that preserves the homographic relation between the input frames. This is achieved with the following components: 1) \textbf{HomE Encoder:} Encodes the input frames into an $n$ dimensional array (where $n$ is the dimension of representation space); 2) \textbf{Vector Neuron module (VN):} Converts an $n$ dimensional 1D vector to an $n\times 3$ dimensional representation; 3) \textbf{Decoder:} Based on the downstream task, takes the $n$ dimensional vector and predicts the corresponding output.}
        \label{fig:model_arch}
    \end{figure}

Any input video $v_i$ from viewpoint $i$ is decoded into a sequence of frames $x_{v_i}$. Representations can be generated by using an encoder network $f_\theta$ and Vector Neuron module (VN) \cite{DBLP:journals/corr/abs-2104-12229} as seen in Figure \ref{fig:model_arch}. The VN module provides a fully-equivariant network, which is used to impose the homographic correspondence between the representations of neighboring views.

% Even though this was originally built to work with Point cloud networks, the Vector neuron module proved useful in our representation learning technique. We impose the homographic correspondence on the $n\times3$ dimensional space (where $n$ is the dimension of the output of the encoder). Instead of using the output of HomE encoder as a collection of scalars, we use the VN module to convert it into a collection of 3D points and that helps us establish the homographic correspondence between the learned points from the different views in the representation vector space.

Let $N_{vp}$ be the total number of viewpoints in the multi-camera setting, so we have $x_{v_1}, x_{v_2}, \ldots, x_{v_{N_{vp}}}$ input frame sequences. All the input videos are decoded into sequences with an equal number of frames, say $T$, and at each time instant, t ($1 \leq t \leq T$), let $x_{t}^{1}, x_{t}^{2}, \ldots, x_{t}^{N_{vp}}$ be the sampled frames. For all pairs of cameras $i$ and $j$,
% (where $j > i$), 
we define the Homography matrices between the two points as $H_{ij}$ and $H_{ji}$, where $H_{ij}$ converts these points in 3D space from viewpoint $i$ to viewpoint $j$ setting, i.e., if $p_i$ and $p_j$ are point correspondences between the two views in homogeneous coordinates, then $p_j \approx H_{ij}.p_i$.
% and $H_{ji}$ converts points from viewpoint $j$ setting to viewpoint $i$ setting. 
For example, in Figure \ref{fig:HomE_Gen_Concept}, we can observe matrices $H_{LC}$ and $H_{CL}$, where $H_{LC}$ converts points from camera $L$ to camera $C$ setting and $H_{CL}$ converts points from camera $C$ to camera $L$ setting.

For each viewpoint $i$, we define $\mathcal{N}(i)$ as all the neighboring viewpoints. In Figure \ref{fig:HomE_Gen_Concept}, cameras $(L, C)$ and $(C, R)$ are neighboring viewpoints while cameras $L$ and $R$ are not. Hence, the neighboring sets for this particular example can be constructed as $\mathcal{N}(L) = \{C\}$, $N(C) = \{L, R\}$, and $\mathcal{N}(R) = \{C\}$. 

Inputs to the model are frames $x_{t}^{1}, x_{t}^{2}, \ldots, x_{t}^{N_{vp}}$ for each $1 \leq t \leq T$, and the representation learning model is trained on minimizing the following loss function, which we refer to as the \textbf{Frobenius loss}: 
\begin{equation}
\mathcal{L}_{\text{Fr}}(x_{v_1},\ldots, x_{v_N}) = \sum_{t=1}^T \sum_{i=1}^{N_{vp}} \sum_{\substack{\mathclap{\substack{j=1\\ j \neq i, j \in \mathcal{N}(i)}}}}^{N_{vp}} \|\mathbf{VN(f(x_{t}^{i}))^T - H_{ji}.VN(f(x_{t}^{j}))^T}\|^2_F
\label{frob_loss}
\end{equation}

The above Frobenius loss facilitates an intuitive approach for implementing the desired equivariance. The square of the Frobenius norm of a matrix is just the sum of squares of all the entries of the matrix. %By minimizing it, we inherently want all entries of the matrix to be as close as possible to zero. 
We want the representations from the input views to preserve the homography operation. The term $VN(f(x_{t}^{i}))^T$ is a representation for View $i$ and $H_{ji}.VN(f(x_{t}^{j}))^T$ transforms points from view representation of $j$ to $i$, and we want the difference to be zero. Hence, we motivate the network by minimizing the squared Frobenius norm of their difference matrix. 

\textbf{\textit{Model architecture:% Representation Learning framework
}} 
As illustrated in Figure \ref{fig:model_arch}, we first define a backbone encoder network $f$ which acts as a frame feature extractor. The image pairs are separately passed into the encoder and we get $n\times1$ dimensional output, where $n$ is the dimension of the vector space we want. The choice for the encoder varies from a simple ResNet model \cite{DBLP:journals/corr/HeZRS15} to SOTA transformer-based models \cite{DBLP:journals/corr/VaswaniSPUJGKP17}. Secondly, we define the Vector Neuron module (VN), which converts an $n\times1$ vector to an $n\times3$ dimensional representation by extending inputs from 1D scalars to 3D vectors. The module consists of layers that preserve the rotational equivariance in 3D. The module enables a simple mapping of SO$(3)$ (3D Rotation group) actions to latent spaces thereby providing a framework for building equivariance in common neural operations. 
% SO$(3)$ is the group of all rotations about the origin of three-dimensional Euclidean space $\mathbf{R^3}$ under the operation of composition.
A deeper dive into the VN module is added later in the section. We enforce the homography preserving criterion through the Frobenius loss, $\mathcal{L}_{\text{Fr}}$, on the outputs from the $VN$ module. In Section \ref{ablation_study}, we perform an ablation study to compare the performance of learned representations on downstream tasks with and without the vector neuron module and validate its significance. Finally, we have the decoder network $g$, which takes the output of the encoder and comes up with the prediction for the downstream task at hand.

In cases where we intend to train the full network jointly with both the representations and the downstream tasks learned together, we jointly optimize them. The overall loss function becomes:
\begin{equation}
    \mathcal{L}_{Total} = \mathcal{L}_{clf}(g(f(x))) + \alpha \mathcal{L}_{Fr}(VN(f(x)))
    \label{Total_loss}
\end{equation} 
where the $\mathcal{L}_{clf}$ is the loss function used to train the supervised training of the downstream task and $\alpha$ is used to balance losses. In the joint optimization case, the encoder not only learns to come up with representations that preserve the homographic property but also keeps the downstream task in mind. Next, We explain the intuition behind using only neighboring views.

\textbf{Intuition Behind Using Only Neighboring Views in the Frobenius Loss:}
% \textbf{\textit{Intuition behind using only neighboring views in Frobenius loss calculation:} } 
% In the methodology section of the paper, we discussed in detail the HomE framework, the Frobenius loss function, and the views involved in learning the required representation. Here we explain why only the neighboring views were used in the loss calculation.
When we calculate the Frobenius loss, for every viewpoint $i$, we only ensured the homographic preservation with its neighbors $N(i)$ and not all the other existing views. This is because we wanted to minimize the redundant information that we force the network to learn. Consider the setting in Figure 1, by forcing the representation of $C$ viewpoint to be related to $L$ and $R$ through $H_{CL}$ and $H_{CR}$, we have already ensured the preservation between views $L$ and $R$ because $H_{LR} \approx H_{CR}.H_{LC}$ (from the properties of homographic matrices). Consider points $p_i,p_j,p_k$ from the three views, by enforcing $p_j \approx H_{LC}.p_i$ and $p_k \approx H_{CR}.p_j$, this gives $p_k \approx H_{CR}.H_{LC}.p_i$, which then entails $p_k \approx H_{LR}.p_i$. Thus in the property preservation between non-neighboring views $L$ and $R$. By enforcing just between the neighboring views we are able to capture the property across all pairs. Enforcing the same thing again is redundant and increases the computations used to learn the representations. A major advantage of HomE is the simple and efficient way in which we can get the representations, and using only the neighbors aids for this cause. 

\textbf{Deep dive into Vector Neuron Module:}
Deng et al.~\cite{DBLP:journals/corr/abs-2104-12229} proposed a simple, lightweight framework to build SO(3) equivariant and invariant point cloud networks. A pivotal component in that framework is a Vector Neuron (VN) representation, extending classical scalar neurons to 3D vectors. Instead of latent vector representations which can be viewed as ordered sequences of scalars, using the Vector neuron representations they can be viewed as (ordered) sequences of 3D vectors.

An appealing property of VN representations is that they remain equivariant to linear layers. Their capabilities do not stop there, they also extend it to a fully-equivariant network with non-linear activations. In particular, the standard neuron-wise activation functions such as ReLU will not commute with a rotation operation. The VN module comes up with a 3D generalization of classical activation functions by implementing them through a learned direction. When applied to a vector neuron, a standard fixed direction ReLU activation would simply truncate the half-plane that points in its opposite direction which gives away the equivariant property that we intend to use. So, they instead dynamically predict an activation direction in a linear data-dependent fashion, and that provides us with the guarantee of equivariance.

Even though this was originally built to work with Point cloud networks, the Vector neuron module proved useful in our representation learning technique. We impose the homographic correspondence on the $n\times3$ dimensional space (where $n$ is the dimension of the output of the encoder). Instead of using the output of HomE encoder as a collection of scalars, we use the VN module to convert it into a collection of 3D points and that helps us establish the homographic correspondence between the learned points from the different views in the representation vector space.

\section{Experiments and Results}
\label{Experiments}
In this section, we dive into the experimental setup and results that demonstrate the effectiveness of our method. The HomE framework is explained in Section \ref{methodology}. HomE can be considered as a frame-level operation making use of the homography matrices between the different views. In order to test the effectiveness of our model, we first take up tasks of increasing complexity starting with simple image classification using the CIFAR$-10$ dataset, ramp it up to action classification from videos of the UCF$101$ dataset, and finally move onto pedestrian action and intent classification on the multi-view video dataset of Stanford-TRI Intent Prediction (STIP). We perform ablation studies to examine the effectiveness of different settings and choices of the model structure. We also do a comparison against other SOTA representation learning frameworks to illustrate HomE capabilities.

\vspace{-3pt}
\subsection{Data}
\label{data}
The CIFAR$-10$ dataset \cite{cifar10} consists of $60,000$ color images of size $32\times32$ distributed between $10$ categories.
% of airplane, automobile, bird, cat, deer, dog, frog, horse, ship, and truck 
% with an equal number of images from each class. 
The train and test split contains $50,000$ training images and $10,000$ test images respectively, maintaining an equal number of samples from each class in both splits. 

The UCF101 dataset \cite{DBLP:journals/corr/abs-1212-0402} is an action classification dataset collected from YouTube, having $13,320$ videos from $101$ action. The videos in the dataset are very diverse in terms of actions, camera motion, object appearance, pose, object scale, cluttered background, and illumination conditions. 

STIP \cite{DBLP:journals/corr/abs-2002-08945} is a dataset of driving scenes recorded in dense urban areas in the United States.
% in California and Michigan, at 8 cities under various weather conditions. 
The dataset contains 923.48 minutes (at 20 fps) 
% 1,108,176 frames in total) 
of driving scenes with high-quality ($1,216 \times 1,936$) recordings. A total of over $350,000$ pedestrian boxes were annotated at 2 fps. 
Each annotated sequence contains video recordings of three cameras simultaneously (left, front, and right). 
% We run two types of experiments on this dataset: (1) Pedestrian action classification by defining the action of every single pedestrian as crossing or not-crossing the street at each time-point; (2) Pedestrian intent prediction, we define each pedestrian's intent using their action in the future (we experiment different settings of 1, 2, or 3 seconds into the future). We follow the experimental setup explained in \cite{DBLP:journals/corr/abs-2002-08945}. 
% They make use of the JRMOT (JackRabbot real-time Multi-Object Tracker) \cite{DBLP:journals/corr/abs-2002-08397} platform to track the pedestrian and interpolate the annotations from 20 fps to 2fps setting.
% \begin{figure}[!]    
%     \begin{subfigure}
%     \centering
%     \quad
%         \includegraphics[width = 0.436 \linewidth]{Images/STIP_sample.png}
%     \end{subfigure}
%     \quad \quad \qaud
%         \begin{subfigure}
%     \centering
%         \includegraphics[width = 0.364 \linewidth]{Images/location_centric.png}
%     \end{subfigure}
%     %\vspace{-8pt}
%     \caption{(a) Sample scenes with left, front, and right camera views from the STIP dataset; (b) Intent prediction under the location-centric setting} 
%     \label{fig:Sample_data}
% \end{figure}

\textbf{\textit{Synthetic Multi-view data creation and Homography estimation:}} 
The homography matrices between different camera views for the STIP dataset were computed using traditional computer vision methods. In the case of CIFAR$-10$ and UCF$101$, the inputs did not originally come from multiple camera view correspondences, hence we synthetically create them using random Homography matrices. A random Homography matrix is created as a random composition of translation, rotation, and scaling operations. The parameters of the transformation matrix were: 1) \textit{Scaling factors along $x$ and $y$-axis} - Random numbers between $[0.5,1.5]$;  2) \textit{Translation along $x$ and $y$-axis} - Random numbers between $[-4,4]$ pixels; 3)\textit{ Rotation in degrees (clockwise)}  - random number between $[-20,20]$ degrees. Two random matrices (to mimic cameras $L$ and $R$) for each dataset were generated and fixed. Using these matrices, we transform the original frames from the dataset and create the multi-view image pairs. The representation learning model now learns to mimic this random matrix correspondence between the viewpoints of the input frames in the learned representation space as well.

\vspace{-3pt}
\subsection{Experimental Setup}
\textbf{\textit{Experimental Details:} }The Decoder network $g$ is an MLP Block consisting of three fully-connected layers with non-linear activations and an output layer. In all of our experiments, the mini-batch size is set to $16$ and the training lasts for a maximum of $200$ epochs.  The learning rate varies from $0.01$ to $0.0001$ in a logarithmic manner for video tasks and between $0.001$ to $0.0001$ for the image classification task. Models with the best performance on the validation data are saved and used for calculating the final accuracy. In the case of fine-tuning using a fixed encoder, we train for a maximum of $50$ epochs with $0.0001$ learning rate. All the numbers reported are averaged values over $10$ runs. We make use of Adam \cite{kingma2017adam} optimizer to train the network. The hyper-parameter in Eq.~\eqref{Total_loss}, $\alpha$, is set to 0.1 to balance the losses. 
% to balance the losses cross-entropy loss and Frobenius loss during joint optimization. 
All models were trained using four NVIDIA Titan GPUs. 

\textbf{\textit{Baseline models}: } 
\label{baseline_models} We compare the performance of our model with other SOTA representation models separately, while through the baseline experiments, we demonstrate that our model also outperforms commonly used supervised learning settings. The following baseline models are used: (a) Supervised setting with pre-trained and fixed encoder and tunable decoder. This is a supervised transfer learning set-up \textbf{(Sup-TL)}; (b)  Supervised setting with no pre-trained weights, 
% for the encoder, 
both the decoder and encoder trained together. This is an ordinary supervised learning setup \textbf{(Sup)}; (c) Pre-trained and fixed HomE encoder, trained on Frobenius loss (Eq.~\eqref{frob_loss}) and tunable decoder fine-tuned on cross-entropy loss. This is also a transfer learning setup, but with HomE encoder \textbf{(HomE-TL)}; (d) Both HomE encoder and decoder are jointly optimized as explained in Eq.~\eqref{Total_loss} \textbf{(HomE-JO)}; (e)  HomE encoder trained on Frobenius loss, while both encoder and decoder fine-tuned on cross-entropy loss \textbf{(HomE)}. 
% Note that acronyms to represent each of the baseline setup is given at the end of point.

\begin{table}[t]
    \centering
    \caption{Accuracy comparison with baseline models on all three datasets. Our model (HomE) with ViT encoder outperforms others across tasks. Best results are typeset in bold. %\textbf{Key:} 
    \textbf{Enc}: Encoder Network; \textbf{PT}: Pre-trained Encoder, Yes or No; \textbf{Fixed}: Encoder weights fixed, Yes or No; \textbf{CIFAR Img clf Acc}: Accuracy (\%) of image classification on CIFAR-10; \textbf{UCF Action clf Acc}:  Accuracy (\%) of action classification on UCF-101; \textbf{STIP Ped Action Acc}: Accuracy (\%) of Pedestrian action classification on STIP;  \textbf{STIP Ped Intent Acc}: Accuracy (\%) of Pedestrian intent prediction (2s $\rightarrow$ 1s) on STIP.}
    
  {\begin{tabular}{lccccccc}
        \toprule
        \textbf{Set Up} & \textbf{Enc} & \textbf{PT}& \textbf{Fixed} & \textbf{\thead{CIFAR Img \\ clf Acc}} & \textbf{\thead{UCF Action \\ clf Acc}} & \textbf{\thead{STIP Ped\\ Action Acc}} & \textbf{\thead{STIP Ped \\ Intent Acc}} \\
        \midrule
        {Sup-TL} & {ResNet} & {\xmark} & {\cmark} &  {88.59} & {86.31} & {79.23} & {76.29} \\ 
        {Sup} & {ResNet} & {\xmark} & {\xmark} &  {92.47}  & {93.70} & {83.74} &  {79.43} \\
        \midrule
        {HomE-TL} & {ResNet} & {\cmark} & {\cmark} & {90.20} & {91.95} & {83.55} &  {79.55} \\
        {HomE-TL} & {ViT} & {\cmark} & {\cmark} & {91.93} & {93.19} & {85.03} &  {83.82} \\
        \midrule
        {HomE-JO} & {ResNet} & {\xmark} & {\xmark} & {89.71} & {89.56} & {81.10} &  {74.81} \\
        {HomE-JO} & {ViT} & {\xmark} & {\xmark} & {92.66} & {91.94} & {83.32} & {80.96} \\
        \midrule
        {HomE (Ours)} & {ResNet} & {\cmark} & {\xmark}  & {92.38} & {94.22} & {86.40} & {81.07} \\
        {HomE (Ours)} & {ViT} & {\cmark} & {\xmark} & \textbf{96.54} & \textbf{96.40} & \textbf{91.25} & \textbf{87.30} \\
        \bottomrule
    \end{tabular}
  }   
  \label{baseline_table} 
\end{table}
\subsection{Results on CIFAR-10 Dataset}
We perform image classification on the synthetic dataset curated from CIFAR-10 
% (refer Sec \ref{data}) 
and compare the performance of different baseline models with two versions of HomE framework using ResNet and Vision Transformer (ViT) \cite{DBLP:journals/corr/abs-2010-11929} models as encoders, and the results are reported in Table \ref{baseline_table}. A thorough comparative study on the performance of different encoder networks is presented in Section \ref{ablation_study}, here we only present the best-performing ViT and ResNet models.

In Table \ref{baseline_table} under the CIFAR column, we can observe that the ViT encoder performs better than the ResNet encoder, giving at least 2\% better performance under all scenarios. We load the ImageNet weights \cite{DBLP:journals/corr/RussakovskyDSKSMHKKBBF14} in the transfer learning setup for the Resnet model and observe that the supervised transfer learning setup (Sup-TL) returns the least accuracy. Using HomE representation based pre-trained weights, which are trained only on 50,000 images (train set of CIFAR), gives more than 4\% improvement in prediction accuracy compared to the ImageNet weights which were originally trained on approximately 1.2 million images, showing that better representations are learned using fewer samples. We also observe that performing joint optimization using both Frobenius and cross-entropy loss comes out to be sub-optimal compared to pre-training and fine-tuning separately. 

\begin{figure}[!]
         \centering
        %(a)
        \includegraphics[width = 0.42\columnwidth]{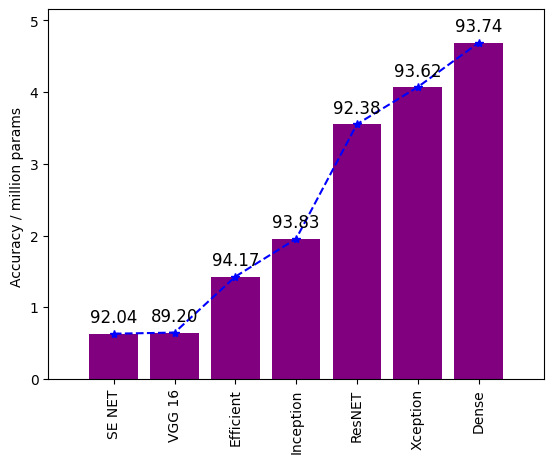} \quad \quad \quad
        %(b)
        \includegraphics[width = 0.48\columnwidth]{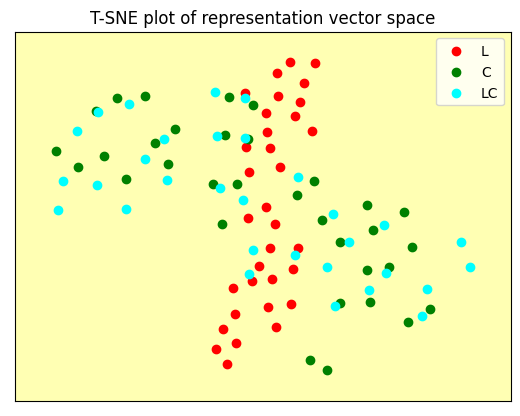}
        \vspace{-8pt}
        \caption{Left: ($\sfrac{\text{Accuracy}}{\text{ \# of million params}}$) for different architectures on CIFAR. Annotations on top of the bars represent the accuracy of predictions from each architecture; Right: Visualizations of HomE Representations using t-SNE, from a subsample of the L and C views from STIP. The L points are transformed to match the C view using the homography matrix $H_{LC}$ between them (denoted by LC).}
        \label{fig:resnet_pic}
    \end{figure}

\textbf{Why ResNet?} %\\
% In order to compare how ResNet performs compared to other commonly used feature extractors,
We compare the performance of ResNet, VGG 16 \cite{simonyan2015deep}, EfficientNet \cite{DBLP:journals/corr/abs-1905-11946}, InceptionNet \cite{DBLP:journals/corr/SzegedyLJSRAEVR14}, XceptionNet \cite{DBLP:journals/corr/Chollet16a}, SENet \cite{DBLP:journals/corr/abs-1709-01507} and DenseNet \cite{DBLP:journals/corr/HuangLW16a} as encoder networks under HomE setting (encoder trained on Frobenius loss, with both encoder and decoder fine-tuned on cross-entropy loss) and the results are presented in Figure \ref{fig:resnet_pic}(a). The reason we start with ResNet for the encoder is twofold: (i) ResNet is the most commonly used image feature extractor and thus helps to establish a point of comparison with existing works. (ii) We observe that ResNet is both simple and efficient (Figure \ref{fig:resnet_pic}(a)), having one of the highest accuracies for every million tunable model parameters after DenseNET and XceptionNET. 
Finally, the numbers from figure \ref{fig:resnet_pic}(a) also demonstrate that ViT outperforms all the $7$ commonly used feature extractors by 3.84\% on average. From image classification, we now step up to evaluate HomE on action classification using the UCF dataset.

%\vspace{-3pt}
\subsection{Results on the UCF-101 Dataset}
We perform action classification on the synthetic dataset curated from UCF-101. $16$ successive frames are sampled from the video clip. Random cropping was conducted to generate input data of size $16 \times 224 \times 224$. 
% We took the central $224$ × $224$
% patch from each frame and ran it through the HomE encoder. 
In addition to random cropping, other augmentations used include random color jittering, Gaussian blur, and flipping.
% The idea here is to decode the input video into sequence of frames and make use of the HomE network to learn and predict the action performed. 
The HomE network here learns from the synthetically generated frames mimicking new camera positions apart from the original frames from the video. The numbers for different baseline settings are reported in Table \ref{baseline_table}.

From Table~\ref{baseline_table} under the UCF column, we can observe the trends between most settings are consistent with that of CIFAR. ViT encoder performs better than the ResNet encoder, this time giving an average of 3.5\% gain in performance. Unlike the case with CIFAR, the HomE setup with ResNet encoder beats the supervised setting on action classification on UCF$-101$. Supervised transfer learning remains to be the worst performer while joint optimization networks are still sub-optimal but now with a bigger gap of 4.46\% drop compared to the HomE setting. This can be attributed to the fact that action classification is a much more challenging task and hence the drop is larger. 

Both CIFAR-10 and UCF-101 datasets had synthetic multi-view pairs generated to make use of the HomE framework, next, we move on to the multi-view video dataset (STIP), and we expect the utility of the HomE framework, built specifically to thrive on multi-view data, to become more visible with larger improvements in performances. 

%\vspace{-3pt}

\subsection{Results on the STIP Dataset}
As humans, when we drive we make decisions unconsciously, such as deciding to stop, speed up, slow down, or move at the same pace. Many of these decisions are made possible by our ability to understand the scene and anticipate the near future. For an autonomous driver, such information is also vital. 
% For example, perceiving pedestrians' intent(s) enables reasoning about the scene and taking appropriate decisions. 
STIP dataset provides a benchmark for this and, here, we use it to evaluate HomE for pedestrian action classification and intent prediction. HomE learns to capture the spatiotemporal context of the surroundings and predict the pedestrians' behaviors. 
% With the help of visual clues and inferring the interdependent interactions among pedestrians and other items in the scene, and we can make a machine to learn it. 
% Our model parses the visual inputs from all the views into the HomE framework to capture the spatiotemporal context of the surroundings and learn to predict the pedestrians' behaviors.

\textbf{\textit{Pedestrian Action classification:}} We intend to predict from the multiple camera views whether the pedestrian has crossed or not. We consider a location-centric setting as described in \cite{DBLP:journals/corr/abs-2002-08945}, where instead of predicting crossing behavior for each pedestrian, we predict the probability that there is someone crossing a designated area.  

\textbf{\textit{Pedestrian Intent prediction:}} Here we go one step further and try to predict the intent of the pedestrian $k$ seconds in the future. The model takes in visual inputs from the last $T$ seconds as past observation and predicts the probability of a pedestrian crossing for the frame $K$ seconds in the future. The baseline computations are done taking $2$s of past observations to predict $1$s into the future $(T =2, K=1)$, denoted by 2s $\rightarrow$ 1s.

% The results are presented under the STIP columns in Table \ref{baseline_table}.
Using a ViT under HomE setup (best performing), we were able to obtain 91.25\% accuracy in cross vs. not-cross prediction and 87.30\% in intent prediction. The difference between the performances of ResNet and ViT keeps increasing with the increase in the complexity of the task. As expected, the difference between the non HomE and HomE settings increases indicating that HomE is indeed more suited for learning better representations with multi-view settings. The previous SOTA using a scene graphs-based model \cite{DBLP:journals/corr/abs-2002-08945} on pedestrian intent prediction was 81.8\% while our model outperforms it by about 6\%. Not just the ViT model, even the simple ResNet encoder (with 81.07\%) learning HomE representations is able to perform as well as a more complex graph-based model \cite{DBLP:journals/corr/abs-2002-08945}.

\textbf{\textit{Predicting further into the future: }}This model for pedestrian intention prediction could be integrated into an autonomous-driving system. For such a system, the ability to predict the pedestrian intent 2s or 3s into the future can be invaluable. Hence, we examine how HomE scales as the time horizon increases, we also compare our performance with SOTA model from \cite{DBLP:journals/corr/abs-2002-08945} in Table \ref{stip_future_pred}.

Our model takes in 2s or 4s of observation and predicts for 1s, 2s, or 3s into the future. The second column only reports the results of HomE with a decoder fine-tuned only on samples from the front camera, and the last column shows the results of all three cameras. In both settings,
% (front only and all three cameras),
predicting for a longer time in the future is seemingly more difficult and thus has lower accuracy and the confidence of predictions at each step decreases monotonically over time. Consistent with \cite{DBLP:journals/corr/abs-2002-08945}, we observe that when using all three cameras, having observed 2s predicts no worse than 4s case. 
We also outperform the SOTA graph model on all prediction lengths. On average, HomE improves predicting the pedestrian intent by 4.7\% when using 2s of observations and 5.4\% using 4s (across both settings), which can be attributed to the quality of representations learned and the capability of the ViT model.

\begin{minipage}[c]{0.5\textwidth}
\captionof{table}{Accuracy (\%) at different input/output setups (input and prediction lengths) %and with SOTA model from \cite{DBLP:journals/corr/abs-2002-08945} on 
for pedestrian intent prediction on STIP.}
\centering
\setlength{\tabcolsep}{2pt}
\vspace{-2pt}
\begin{tabular}{ccccc}
        \toprule
        \multirow{2}{*}{Setup} %& \multicolumn{4}{c|} {\textbf{Accuracy (\%)}} \\ 
        & \multicolumn{2}{c}{\textbf{Front camera}}  & \multicolumn{2}{c}{\textbf{All cameras}} \\ 
        \cmidrule(lr){2-3}\cmidrule(lr){4-5}
        ~ &
        \textbf{HomE}  & \textbf{Graph} \cite{DBLP:journals/corr/abs-2002-08945}  & \textbf{HomE}  & \textbf{Graph} \cite{DBLP:journals/corr/abs-2002-08945} \\ 
        \midrule
        2s $\rightarrow$ 1s & 83.82 & 78.68 & 87.30 & 81.20 \\ 
        2s $\rightarrow$ 2s & 81.40 & 78.09 & 86.21 & 80.49 \\ 
        2s $\rightarrow$ 3s & 80.63 & 78.16 & 86.04 & 80.77 \\ 
        4s $\rightarrow$ 1s & 86.37 & 80.36 & 88.67 & 81.53 \\ 
        4s $\rightarrow$ 2s & 84.28 & 80.06 & 86.80  & 81.73 \\ 
        4s $\rightarrow$ 3s & 84.55 & 80.32 & 85.12 & 79.62 \\ 
        \bottomrule
    \end{tabular}
\label{stip_future_pred}
\end{minipage}
\qquad 
\begin{minipage}[c]{0.4\textwidth}
\captionof{table}{Comparison with SOTA representation learning methods.}
\centering
\vspace{-2pt}
\begin{tabular}{lcc}
        \toprule
        \textbf{Model} & \textbf{\thead{STIP Ped \\ Action Acc}} & \textbf{\thead{STIP Ped \\ Intent Acc}} \\ \midrule
        InsDis \cite{DBLP:journals/corr/abs-1805-01978} & 86.48 & 72.21  \\
        MoCo \cite{DBLP:journals/corr/abs-1911-05722} & 83.35 &  76.52 \\
        CMC \cite{DBLP:journals/corr/abs-1906-05849} & 88.72 &   79.66 \\ 
        TCN \cite{8014803} & 87.10 & 80.89  \\
        HomE (Ours) & \textbf{ 91.25} & \textbf{87.30}  \\
        \bottomrule
    \end{tabular}
\label{SOTA_comparison}
\end{minipage}

\textbf{\textit{Comparison with SOTA representation learning models:}} We compare the performance of HomE (our model) with other SOTA models on pedestrian intent and action classification on STIP dataset in Table \ref{SOTA_comparison}. Time-Contrastive Network (TCN) \cite{8014803} tries to push frames from the same time from all views to align, while Contrastive multi-view coding (CMC) \cite{DBLP:journals/corr/abs-1906-05849} tries to align all the views in the representation space. We also compare with non-multi-view techniques: Momentum Contrast (MoCo) \cite{DBLP:journals/corr/abs-1911-05722} learns a visual representation encoder by matching an encoded query to a dictionary of encoded keys using a contrastive loss and Instance-level Discrimination (InsDis) \cite{DBLP:journals/corr/abs-1805-01978} which captures visual similarity from the output of a neural network and group samples in positive and negative pairs to perform contrastive learning. 

From Table \ref{SOTA_comparison} we observe that our model outperforms both the SOTA multi-view and non-multi-view techniques. The average improvements over multi-view techniques are 3.3\% and 8.5\%, respectively, for pedestrian action classification and intent prediction, while the same versus non-multi views are 6.3\% and 12.9\%.
As expected, MoCo and InsDis had the least numbers and this may be because they were not trained to handle the multi-view nature of STIP dataset. Interestingly, the CMC model performs better than TSN in the action classification task but the same trend does not hold true for intent prediction. The improvement we get over the existing multi-view techniques can be pointed toward the benefits of using non-aligned but related representations for multiple views.

\textbf{\textbf{\textit{Visualizing the Learned Vector Space}:}}
Finally, we visualize the vector representation space learned using the HomE framework in Figure \ref{fig:resnet_pic}(b). We plot the representations of input frames from L and C cameras. Simultaneously, we also plot the representations of points transformed from L to C view via $H_{LC}$ to demonstrate the homographic correspondence as postulated throughout this work. The $LC$ points do not exactly coincide with $C$ points, this shows both that the representation learning model is not overfitted and the fact that after pre-training, the framework is fine-tuned on cross-entropy loss and we expect that to push to points to suit the downstream task better than obtaining perfect matches.

%\vspace{-4pt}
\subsection{Ablation study}
The HomE framework, as seen in Figure \ref{fig:model_arch}, consists of a HomE encoder and Vector Neuron (VN) module. In this section, we still study the effect of the choice of encoder on model performance and understand the benefits of using the Vector Neuron module.

\begin{figure}[]
\CenterFloatBoxes
\begin{floatrow}
\ffigbox
  { \includegraphics[width = 0.80 \columnwidth]{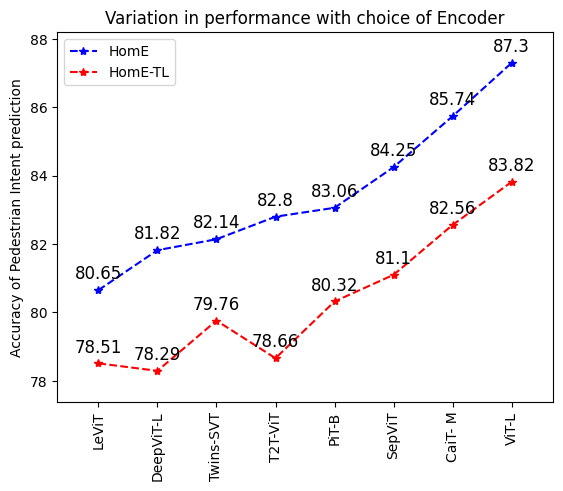}\vspace{-5pt}}
  {\caption{Ablation study on Encoder network: Comparisons using different encoders.}
  \label{fig:ViT_ablation}}
\killfloatstyle
\ttabbox
  {\begin{tabular}{lccc}
        \toprule
        \textbf{Dataset} & \textbf{Enc} &\textbf{\thead{Acc\\ with VN}} & \textbf{\thead{Acc\\ w/o VN}} \\ 
        \midrule
        CIFAR-10 & ResNet &  92.38 & 88.4 \\
        CIFAR-10 & ViT  &  96.54 & 92.1 \\
        UCF-101  & ResNet & 94.22 & 89.7 \\
        UCF-101  & ViT & 96.40 & 90.9 \\
        STIP (Intent) & ResNet & 81.07 & 78.7 \\
        STIP (Intent) & ViT & 87.30 & 80.2 \\
        \bottomrule
    \end{tabular}
  }
  {\caption{Ablation study on VN module: Comparisons with and without the VN module.}
  \label{tab:VN_ablation}}
\end{floatrow}
\end{figure}

\textbf{\textit{Ablation study on Encoder Network:}}
The encoder network plays a pivotal role in learning good representations. It is the output of the encoder network that is fed into the Vector neuron module to minimize Frobenius loss and also the one passed as input to the decoder block to make predictions. Table \ref{baseline_table} clearly shows the gains we could get by using a ViT encoder instead of ResNet. Similarly, from Figure \ref{fig:resnet_pic}(a), we have also seen that the ViT outperforms the common image feature extractors. We now extend these encoder choices to different variants of the image transformers: DeepViT \cite{DBLP:journals/corr/abs-2103-11886}, CaiT \cite{DBLP:journals/corr/abs-2103-17239},  PiT \cite{DBLP:journals/corr/abs-2103-16302}, LeViT \cite{DBLP:journals/corr/abs-2104-01136}, SepViT \cite{li2022sepvit}, T2T-ViT \cite{DBLP:journals/corr/abs-2101-11986} and  Twins-SVT \cite{DBLP:journals/corr/abs-2104-13840}. We compare their performances on pedestrian intent prediction on STIP dataset predicting 1s into the future, and the numbers are presented in Figure \ref{fig:ViT_ablation}. 

We can observe that the ViT model outperforms the other choices by at least 4\%. We observe similar trends with encoder choices both in HomE and HomE-TL settings. Note that HomE learning technique proposed by us, in no way forces ViT as the encoder network, 
% all we do with HomE is show that learning representations preserving Homographic correspondence helps in learning better representations and 
the choice of the best encoder may vary from task to task, but the ViT can be expected to give good performance in general.

\label{ablation_study}

\textbf{\textit{Ablation study on Vector Neuron module:}}
The other component of HomE framework is the Vector neuron module. This module takes a $n\times1$ vector and converts it into $n\times3$ by extending 1D scalars into 3D vectors. Instead of using a Vector Neuron module, we replace it with three separate projection heads which will be concatenated to give the $n\times3$ output. This representation is not guaranteed to have equivariant property, unlike the output of the Vector Neuron module. In Table \ref{tab:VN_ablation}, we compare the performances of HomE framework, with and without the VN module on all three datasets. 
We observe that the average dip across datasets when we do not use VN module is approximately 5\%. The drop in performance is more severe for ViT encoders. In terms of datasets, the drop is maximum for pedestrian intent prediction on STIP dataset and this is expected due to its multi-view nature.
%\vspace{-4pt}
\section{Conclusion}
\label{conclusion}
In this paper, we have proposed Homography equivariant representation learning (HomE), a simple and effective framework for learning representations for multi-view data. Experiments comparing HomE with other representation learning techniques show how our model outperforms the SOTA models. An extensive ablation study showed the effectiveness of the Vector Neuron module and the ability of ViT as a HomE encoder. 
% Our proposed HomE is flexible enough and can be easily extended to any downstream task of choice. 
A limitation of this study is that we only test the algorithm on three cameras. As a direction for future work, one can work towards understanding whether or not the representation quality improves as the number of views increases. With HomE, we do not just focus on lifting the baselines to a new level but also intended to show the utility of an intuitive and simple technique to learn multi-view representations. 

\section*{Acknowledgment}
This research was partially supported by the Toyota Research Institute (TRI). This article solely reflects the opinions and conclusions of its authors and not TRI or any other Toyota entity.

%%%%%%%%%%%%%%%%%%%%%%%%%%%%%%%%%%%%%%%%%%%%%%%%%%%%%%%%%%%%

\bibliographystyle{acl_natbib}
\bibliography{references}

\begin{thebibliography}{78}
\expandafter\ifx\csname natexlab\endcsname\relax\def\natexlab#1{#1}\fi

\bibitem[{Abu{-}El{-}Haija et~al.(2016)Abu{-}El{-}Haija, Kothari, Lee, Natsev,
  Toderici, Varadarajan, and
  Vijayanarasimhan}]{DBLP:journals/corr/Abu-El-HaijaKLN16}
Sami Abu{-}El{-}Haija, Nisarg Kothari, Joonseok Lee, Paul Natsev, George
  Toderici, Balakrishnan Varadarajan, and Sudheendra Vijayanarasimhan. 2016.
\newblock \href {http://arxiv.org/abs/1609.08675} {Youtube-8m: {A} large-scale
  video classification benchmark}.
\newblock \emph{CoRR}, abs/1609.08675.

\bibitem[{Alwassel et~al.(2019)Alwassel, Mahajan, Torresani, Ghanem, and
  Tran}]{DBLP:journals/corr/abs-1911-12667}
Humam Alwassel, Dhruv Mahajan, Lorenzo Torresani, Bernard Ghanem, and Du~Tran.
  2019.
\newblock \href {http://arxiv.org/abs/1911.12667} {Self-supervised learning by
  cross-modal audio-video clustering}.
\newblock \emph{CoRR}, abs/1911.12667.

\bibitem[{Asano et~al.(2019)Asano, Rupprecht, and
  Vedaldi}]{DBLP:journals/corr/abs-1911-05371}
Yuki~Markus Asano, Christian Rupprecht, and Andrea Vedaldi. 2019.
\newblock \href {http://arxiv.org/abs/1911.05371} {Self-labelling via
  simultaneous clustering and representation learning}.
\newblock \emph{CoRR}, abs/1911.05371.

\bibitem[{Bachman et~al.(2019)Bachman, Hjelm, and
  Buchwalter}]{DBLP:journals/corr/abs-1906-00910}
Philip Bachman, R.~Devon Hjelm, and William Buchwalter. 2019.
\newblock \href {http://arxiv.org/abs/1906.00910} {Learning representations by
  maximizing mutual information across views}.
\newblock \emph{CoRR}, abs/1906.00910.

\bibitem[{Bautista et~al.(2016)Bautista, Sanakoyeu, Sutter, and
  Ommer}]{DBLP:journals/corr/BautistaSSO16}
Miguel~{\'{A}}ngel Bautista, Artsiom Sanakoyeu, Ekaterina Sutter, and
  Bj{\"{o}}rn Ommer. 2016.
\newblock \href {http://arxiv.org/abs/1608.08792} {Cliquecnn: Deep unsupervised
  exemplar learning}.
\newblock \emph{CoRR}, abs/1608.08792.

\bibitem[{Bengio and Bergstra(2009)}]{NIPS2009_043c3d7e}
Yoshua Bengio and James Bergstra. 2009.
\newblock \href
  {https://proceedings.neurips.cc/paper_files/paper/2009/file/043c3d7e489c69b48737cc0c92d0f3a2-Paper.pdf}
  {Slow, decorrelated features for pretraining complex cell-like networks}.
\newblock In \emph{Advances in Neural Information Processing Systems},
  volume~22. Curran Associates, Inc.

\bibitem[{Bilen et~al.(2016)Bilen, Fernando, Gavves, and
  Vedaldi}]{DBLP:journals/corr/BilenFGV16}
Hakan Bilen, Basura Fernando, Efstratios Gavves, and Andrea Vedaldi. 2016.
\newblock \href {http://arxiv.org/abs/1612.00738} {Action recognition with
  dynamic image networks}.
\newblock \emph{CoRR}, abs/1612.00738.

\bibitem[{Caron et~al.(2018)Caron, Bojanowski, Joulin, and
  Douze}]{DBLP:journals/corr/abs-1807-05520}
Mathilde Caron, Piotr Bojanowski, Armand Joulin, and Matthijs Douze. 2018.
\newblock \href {http://arxiv.org/abs/1807.05520} {Deep clustering for
  unsupervised learning of visual features}.
\newblock \emph{CoRR}, abs/1807.05520.

\bibitem[{Caron et~al.(2020)Caron, Misra, Mairal, Goyal, Bojanowski, and
  Joulin}]{DBLP:journals/corr/abs-2006-09882}
Mathilde Caron, Ishan Misra, Julien Mairal, Priya Goyal, Piotr Bojanowski, and
  Armand Joulin. 2020.
\newblock \href {http://arxiv.org/abs/2006.09882} {Unsupervised learning of
  visual features by contrasting cluster assignments}.
\newblock \emph{CoRR}, abs/2006.09882.

\bibitem[{Chen et~al.(2020{\natexlab{a}})Chen, Kornblith, Norouzi, and
  Hinton}]{DBLP:conf/icml/ChenK0H20}
Ting Chen, Simon Kornblith, Mohammad Norouzi, and Geoffrey~E. Hinton.
  2020{\natexlab{a}}.
\newblock \href {http://proceedings.mlr.press/v119/chen20j.html} {A simple
  framework for contrastive learning of visual representations}.
\newblock In \emph{Proceedings of the 37th International Conference on Machine
  Learning, {ICML} 2020, 13-18 July 2020, Virtual Event}, volume 119 of
  \emph{Proceedings of Machine Learning Research}, pages 1597--1607. {PMLR}.

\bibitem[{Chen et~al.(2020{\natexlab{b}})Chen, Fan, Girshick, and
  He}]{DBLP:journals/corr/abs-2003-04297}
Xinlei Chen, Haoqi Fan, Ross~B. Girshick, and Kaiming He. 2020{\natexlab{b}}.
\newblock \href {http://arxiv.org/abs/2003.04297} {Improved baselines with
  momentum contrastive learning}.
\newblock \emph{CoRR}, abs/2003.04297.

\bibitem[{Cho et~al.(2020)Cho, Kim, Chang, and
  Hwang}]{DBLP:journals/corr/abs-2003-02692}
Hyeon Cho, Taehoon Kim, Hyung~Jin Chang, and Wonjun Hwang. 2020.
\newblock \href {http://arxiv.org/abs/2003.02692} {Self-supervised
  spatio-temporal representation learning using variable playback speed
  prediction}.
\newblock \emph{CoRR}, abs/2003.02692.

\bibitem[{Chollet(2016)}]{DBLP:journals/corr/Chollet16a}
Fran{\c{c}}ois Chollet. 2016.
\newblock \href {http://arxiv.org/abs/1610.02357} {Xception: Deep learning with
  depthwise separable convolutions}.
\newblock \emph{CoRR}, abs/1610.02357.

\bibitem[{Chu et~al.(2021)Chu, Tian, Wang, Zhang, Ren, Wei, Xia, and
  Shen}]{DBLP:journals/corr/abs-2104-13840}
Xiangxiang Chu, Zhi Tian, Yuqing Wang, Bo~Zhang, Haibing Ren, Xiaolin Wei,
  Huaxia Xia, and Chunhua Shen. 2021.
\newblock \href {http://arxiv.org/abs/2104.13840} {Twins: Revisiting spatial
  attention design in vision transformers}.
\newblock \emph{CoRR}, abs/2104.13840.

\bibitem[{Deng et~al.(2021)Deng, Litany, Duan, Poulenard, Tagliasacchi, and
  Guibas}]{DBLP:journals/corr/abs-2104-12229}
Congyue Deng, Or~Litany, Yueqi Duan, Adrien Poulenard, Andrea Tagliasacchi, and
  Leonidas~J. Guibas. 2021.
\newblock \href {http://arxiv.org/abs/2104.12229} {Vector neurons: {A} general
  framework for so(3)-equivariant networks}.
\newblock \emph{CoRR}, abs/2104.12229.

\bibitem[{Doersch et~al.(2015)Doersch, Gupta, and
  Efros}]{DBLP:journals/corr/DoerschGE15}
Carl Doersch, Abhinav Gupta, and Alexei~A. Efros. 2015.
\newblock \href {http://arxiv.org/abs/1505.05192} {Unsupervised visual
  representation learning by context prediction}.
\newblock \emph{CoRR}, abs/1505.05192.

\bibitem[{Dosovitskiy et~al.(2020)Dosovitskiy, Beyer, Kolesnikov, Weissenborn,
  Zhai, Unterthiner, Dehghani, Minderer, Heigold, Gelly, Uszkoreit, and
  Houlsby}]{DBLP:journals/corr/abs-2010-11929}
Alexey Dosovitskiy, Lucas Beyer, Alexander Kolesnikov, Dirk Weissenborn,
  Xiaohua Zhai, Thomas Unterthiner, Mostafa Dehghani, Matthias Minderer, Georg
  Heigold, Sylvain Gelly, Jakob Uszkoreit, and Neil Houlsby. 2020.
\newblock \href {http://arxiv.org/abs/2010.11929} {An image is worth 16x16
  words: Transformers for image recognition at scale}.
\newblock \emph{CoRR}, abs/2010.11929.

\bibitem[{Faghri et~al.(2017)Faghri, Fleet, Kiros, and
  Fidler}]{DBLP:journals/corr/FaghriFKF17}
Fartash Faghri, David~J. Fleet, Jamie~Ryan Kiros, and Sanja Fidler. 2017.
\newblock \href {http://arxiv.org/abs/1707.05612} {{VSE++:} improved
  visual-semantic embeddings}.
\newblock \emph{CoRR}, abs/1707.05612.

\bibitem[{Fernando et~al.(2016)Fernando, Bilen, Gavves, and
  Gould}]{DBLP:journals/corr/FernandoBGG16}
Basura Fernando, Hakan Bilen, Efstratios Gavves, and Stephen Gould. 2016.
\newblock \href {http://arxiv.org/abs/1611.06646} {Self-supervised video
  representation learning with odd-one-out networks}.
\newblock \emph{CoRR}, abs/1611.06646.

\bibitem[{Gidaris et~al.(2020)Gidaris, Bursuc, Komodakis, P{\'{e}}rez, and
  Cord}]{DBLP:journals/corr/abs-2002-12247}
Spyros Gidaris, Andrei Bursuc, Nikos Komodakis, Patrick P{\'{e}}rez, and
  Matthieu Cord. 2020.
\newblock \href {http://arxiv.org/abs/2002.12247} {Learning representations by
  predicting bags of visual words}.
\newblock \emph{CoRR}, abs/2002.12247.

\bibitem[{Gidaris et~al.(2018)Gidaris, Singh, and
  Komodakis}]{DBLP:journals/corr/abs-1803-07728}
Spyros Gidaris, Praveer Singh, and Nikos Komodakis. 2018.
\newblock \href {http://arxiv.org/abs/1803.07728} {Unsupervised representation
  learning by predicting image rotations}.
\newblock \emph{CoRR}, abs/1803.07728.

\bibitem[{Graham et~al.(2021)Graham, El{-}Nouby, Touvron, Stock, Joulin,
  J{\'{e}}gou, and Douze}]{DBLP:journals/corr/abs-2104-01136}
Benjamin Graham, Alaaeldin El{-}Nouby, Hugo Touvron, Pierre Stock, Armand
  Joulin, Herv{\'{e}} J{\'{e}}gou, and Matthijs Douze. 2021.
\newblock \href {http://arxiv.org/abs/2104.01136} {Levit: a vision transformer
  in convnet's clothing for faster inference}.
\newblock \emph{CoRR}, abs/2104.01136.

\bibitem[{Grill et~al.(2020)Grill, Strub, Altch{\'{e}}, Tallec, Richemond,
  Buchatskaya, Doersch, Pires, Guo, Azar, Piot, Kavukcuoglu, Munos, and
  Valko}]{DBLP:journals/corr/abs-2006-07733}
Jean{-}Bastien Grill, Florian Strub, Florent Altch{\'{e}}, Corentin Tallec,
  Pierre~H. Richemond, Elena Buchatskaya, Carl Doersch, Bernardo~{\'{A}}vila
  Pires, Zhaohan~Daniel Guo, Mohammad~Gheshlaghi Azar, Bilal Piot, Koray
  Kavukcuoglu, R{\'{e}}mi Munos, and Michal Valko. 2020.
\newblock \href {http://arxiv.org/abs/2006.07733} {Bootstrap your own latent:
  {A} new approach to self-supervised learning}.
\newblock \emph{CoRR}, abs/2006.07733.

\bibitem[{Gutmann and Hyvärinen(2010)}]{pmlr-v9-gutmann10a}
Michael Gutmann and Aapo Hyvärinen. 2010.
\newblock \href {https://proceedings.mlr.press/v9/gutmann10a.html}
  {Noise-contrastive estimation: A new estimation principle for unnormalized
  statistical models}.
\newblock In \emph{Proceedings of the Thirteenth International Conference on
  Artificial Intelligence and Statistics}, volume~9 of \emph{Proceedings of
  Machine Learning Research}, pages 297--304, Chia Laguna Resort, Sardinia,
  Italy. PMLR.

\bibitem[{Han et~al.(2019)Han, Xie, and
  Zisserman}]{DBLP:journals/corr/abs-1909-04656}
Tengda Han, Weidi Xie, and Andrew Zisserman. 2019.
\newblock \href {http://arxiv.org/abs/1909.04656} {Video representation
  learning by dense predictive coding}.
\newblock \emph{CoRR}, abs/1909.04656.

\bibitem[{Han et~al.(2020{\natexlab{a}})Han, Xie, and
  Zisserman}]{DBLP:journals/corr/abs-2008-01065}
Tengda Han, Weidi Xie, and Andrew Zisserman. 2020{\natexlab{a}}.
\newblock \href {http://arxiv.org/abs/2008.01065} {Memory-augmented dense
  predictive coding for video representation learning}.
\newblock \emph{CoRR}, abs/2008.01065.

\bibitem[{Han et~al.(2020{\natexlab{b}})Han, Xie, and
  Zisserman}]{DBLP:journals/corr/abs-2010-09709}
Tengda Han, Weidi Xie, and Andrew Zisserman. 2020{\natexlab{b}}.
\newblock \href {http://arxiv.org/abs/2010.09709} {Self-supervised co-training
  for video representation learning}.
\newblock \emph{CoRR}, abs/2010.09709.

\bibitem[{He et~al.(2019)He, Fan, Wu, Xie, and
  Girshick}]{DBLP:journals/corr/abs-1911-05722}
Kaiming He, Haoqi Fan, Yuxin Wu, Saining Xie, and Ross~B. Girshick. 2019.
\newblock \href {http://arxiv.org/abs/1911.05722} {Momentum contrast for
  unsupervised visual representation learning}.
\newblock \emph{CoRR}, abs/1911.05722.

\bibitem[{He et~al.(2015)He, Zhang, Ren, and Sun}]{DBLP:journals/corr/HeZRS15}
Kaiming He, Xiangyu Zhang, Shaoqing Ren, and Jian Sun. 2015.
\newblock \href {http://arxiv.org/abs/1512.03385} {Deep residual learning for
  image recognition}.
\newblock \emph{CoRR}, abs/1512.03385.

\bibitem[{Heo et~al.(2021)Heo, Yun, Han, Chun, Choe, and
  Oh}]{DBLP:journals/corr/abs-2103-16302}
Byeongho Heo, Sangdoo Yun, Dongyoon Han, Sanghyuk Chun, Junsuk Choe, and
  Seong~Joon Oh. 2021.
\newblock \href {http://arxiv.org/abs/2103.16302} {Rethinking spatial
  dimensions of vision transformers}.
\newblock \emph{CoRR}, abs/2103.16302.

\bibitem[{Hjelm and Bachman(2020)}]{DBLP:journals/corr/abs-2007-13278}
R.~Devon Hjelm and Philip Bachman. 2020.
\newblock \href {http://arxiv.org/abs/2007.13278} {Representation learning with
  video deep infomax}.
\newblock \emph{CoRR}, abs/2007.13278.

\bibitem[{Hjelm et~al.(2019)Hjelm, Fedorov, Lavoie-Marchildon, Grewal, Bachman,
  Trischler, and Bengio}]{hjelm2019learning}
R~Devon Hjelm, Alex Fedorov, Samuel Lavoie-Marchildon, Karan Grewal, Phil
  Bachman, Adam Trischler, and Yoshua Bengio. 2019.
\newblock \href {http://arxiv.org/abs/1808.06670} {Learning deep
  representations by mutual information estimation and maximization}.

\bibitem[{Hu et~al.(2017)Hu, Shen, and Sun}]{DBLP:journals/corr/abs-1709-01507}
Jie Hu, Li~Shen, and Gang Sun. 2017.
\newblock \href {http://arxiv.org/abs/1709.01507} {Squeeze-and-excitation
  networks}.
\newblock \emph{CoRR}, abs/1709.01507.

\bibitem[{Huang et~al.(2016)Huang, Liu, and
  Weinberger}]{DBLP:journals/corr/HuangLW16a}
Gao Huang, Zhuang Liu, and Kilian~Q. Weinberger. 2016.
\newblock \href {http://arxiv.org/abs/1608.06993} {Densely connected
  convolutional networks}.
\newblock \emph{CoRR}, abs/1608.06993.

\bibitem[{Huang et~al.(2019)Huang, Dong, Gong, and
  Zhu}]{DBLP:journals/corr/abs-1904-11567}
Jiabo Huang, Qi~Dong, Shaogang Gong, and Xiatian Zhu. 2019.
\newblock \href {http://arxiv.org/abs/1904.11567} {Unsupervised deep learning
  by neighbourhood discovery}.
\newblock \emph{CoRR}, abs/1904.11567.

\bibitem[{Jenni et~al.(2020)Jenni, Meishvili, and
  Favaro}]{DBLP:journals/corr/abs-2007-10730}
Simon Jenni, Givi Meishvili, and Paolo Favaro. 2020.
\newblock \href {http://arxiv.org/abs/2007.10730} {Video representation
  learning by recognizing temporal transformations}.
\newblock \emph{CoRR}, abs/2007.10730.

\bibitem[{Jing and Tian(2018)}]{DBLP:journals/corr/abs-1811-11387}
Longlong Jing and Yingli Tian. 2018.
\newblock \href {http://arxiv.org/abs/1811.11387} {Self-supervised
  spatiotemporal feature learning by video geometric transformations}.
\newblock \emph{CoRR}, abs/1811.11387.

\bibitem[{Karpathy et~al.(2014)Karpathy, Toderici, Shetty, Leung, Sukthankar,
  and Fei-Fei}]{6909619}
Andrej Karpathy, George Toderici, Sanketh Shetty, Thomas Leung, Rahul
  Sukthankar, and Li~Fei-Fei. 2014.
\newblock \href {https://doi.org/10.1109/CVPR.2014.223} {Large-scale video
  classification with convolutional neural networks}.
\newblock In \emph{2014 IEEE Conference on Computer Vision and Pattern
  Recognition}, pages 1725--1732.

\bibitem[{Kim et~al.(2018)Kim, Cho, and
  Kweon}]{DBLP:journals/corr/abs-1811-09795}
Dahun Kim, Donghyeon Cho, and In~So Kweon. 2018.
\newblock \href {http://arxiv.org/abs/1811.09795} {Self-supervised video
  representation learning with space-time cubic puzzles}.
\newblock \emph{CoRR}, abs/1811.09795.

\bibitem[{Kingma and Ba(2017)}]{kingma2017adam}
Diederik~P. Kingma and Jimmy Ba. 2017.
\newblock \href {http://arxiv.org/abs/1412.6980} {Adam: A method for stochastic
  optimization}.

\bibitem[{Krizhevsky et~al.(2009)Krizhevsky, Nair, and Hinton}]{cifar10}
Alex Krizhevsky, Vinod Nair, and Geoffrey Hinton. 2009.
\newblock \href {http://www.cs.toronto.edu/~kriz/cifar.html} {Cifar-10
  (canadian institute for advanced research)}.
\newblock \emph{-}.

\bibitem[{Kuehne et~al.(2011)Kuehne, Jhuang, Garrote, Poggio, and
  Serre}]{6126543}
H.~Kuehne, H.~Jhuang, E.~Garrote, T.~Poggio, and T.~Serre. 2011.
\newblock \href {https://doi.org/10.1109/ICCV.2011.6126543} {Hmdb: A large
  video database for human motion recognition}.
\newblock In \emph{2011 International Conference on Computer Vision}, pages
  2556--2563.

\bibitem[{Lee et~al.(2017)Lee, Huang, Singh, and
  Yang}]{DBLP:journals/corr/abs-1708-01246}
Hsin{-}Ying Lee, Jia{-}Bin Huang, Maneesh Singh, and Ming{-}Hsuan Yang. 2017.
\newblock \href {http://arxiv.org/abs/1708.01246} {Unsupervised representation
  learning by sorting sequences}.
\newblock \emph{CoRR}, abs/1708.01246.

\bibitem[{Li et~al.(2022)Li, Wang, Xia, Wu, Xiao, Zheng, and
  Wen}]{li2022sepvit}
Wei Li, Xing Wang, Xin Xia, Jie Wu, Xuefeng Xiao, Min Zheng, and Shiping Wen.
  2022.
\newblock \href {http://arxiv.org/abs/2203.15380} {Sepvit: Separable vision
  transformer}.

\bibitem[{Liu et~al.(2020)Liu, Adeli, Cao, Lee, Shenoi, Gaidon, and
  Niebles}]{DBLP:journals/corr/abs-2002-08945}
Bingbin Liu, Ehsan Adeli, Zhangjie Cao, Kuan{-}Hui Lee, Abhijeet Shenoi, Adrien
  Gaidon, and Juan~Carlos Niebles. 2020.
\newblock \href {http://arxiv.org/abs/2002.08945} {Spatiotemporal relationship
  reasoning for pedestrian intent prediction}.
\newblock \emph{CoRR}, abs/2002.08945.

\bibitem[{Mathieu et~al.(2016)Mathieu, Couprie, and
  LeCun}]{DBLP:journals/corr/MathieuCL15}
Micha{\"{e}}l Mathieu, Camille Couprie, and Yann LeCun. 2016.
\newblock \href {http://arxiv.org/abs/1511.05440} {Deep multi-scale video
  prediction beyond mean square error}.
\newblock In \emph{4th International Conference on Learning Representations,
  {ICLR} 2016, San Juan, Puerto Rico, May 2-4, 2016, Conference Track
  Proceedings}.

\bibitem[{Misra et~al.(2016)Misra, Zitnick, and
  Hebert}]{DBLP:journals/corr/MisraZH16}
Ishan Misra, C.~Lawrence Zitnick, and Martial Hebert. 2016.
\newblock \href {http://arxiv.org/abs/1603.08561} {Unsupervised learning using
  sequential verification for action recognition}.
\newblock \emph{CoRR}, abs/1603.08561.

\bibitem[{Mobahi et~al.(2009)Mobahi, Collobert, and
  Weston}]{10.1145/1553374.1553469}
Hossein Mobahi, Ronan Collobert, and Jason Weston. 2009.
\newblock \href {https://doi.org/10.1145/1553374.1553469} {Deep learning from
  temporal coherence in video}.
\newblock In \emph{Proceedings of the 26th Annual International Conference on
  Machine Learning}, ICML '09, page 737–744, New York, NY, USA. Association
  for Computing Machinery.

\bibitem[{Noroozi and Favaro(2016)}]{DBLP:journals/corr/NorooziF16}
Mehdi Noroozi and Paolo Favaro. 2016.
\newblock \href {http://arxiv.org/abs/1603.09246} {Unsupervised learning of
  visual representations by solving jigsaw puzzles}.
\newblock \emph{CoRR}, abs/1603.09246.

\bibitem[{van~den Oord et~al.(2018)van~den Oord, Li, and
  Vinyals}]{DBLP:journals/corr/abs-1807-03748}
A{\"{a}}ron van~den Oord, Yazhe Li, and Oriol Vinyals. 2018.
\newblock \href {http://arxiv.org/abs/1807.03748} {Representation learning with
  contrastive predictive coding}.
\newblock \emph{CoRR}, abs/1807.03748.

\bibitem[{Pathak et~al.(2016)Pathak, Kr{\"{a}}henb{\"{u}}hl, Donahue, Darrell,
  and Efros}]{DBLP:journals/corr/PathakKDDE16}
Deepak Pathak, Philipp Kr{\"{a}}henb{\"{u}}hl, Jeff Donahue, Trevor Darrell,
  and Alexei~A. Efros. 2016.
\newblock \href {http://arxiv.org/abs/1604.07379} {Context encoders: Feature
  learning by inpainting}.
\newblock \emph{CoRR}, abs/1604.07379.

\bibitem[{Patraucean et~al.(2015)Patraucean, Handa, and
  Cipolla}]{DBLP:journals/corr/PatrauceanHC15}
Viorica Patraucean, Ankur Handa, and Roberto Cipolla. 2015.
\newblock \href {http://arxiv.org/abs/1511.06309} {Spatio-temporal video
  autoencoder with differentiable memory}.
\newblock \emph{CoRR}, abs/1511.06309.

\bibitem[{Qian et~al.(2020)Qian, Meng, Gong, Yang, Wang, Belongie, and
  Cui}]{DBLP:journals/corr/abs-2008-03800}
Rui Qian, Tianjian Meng, Boqing Gong, Ming{-}Hsuan Yang, Huisheng Wang,
  Serge~J. Belongie, and Yin Cui. 2020.
\newblock \href {http://arxiv.org/abs/2008.03800} {Spatiotemporal contrastive
  video representation learning}.
\newblock \emph{CoRR}, abs/2008.03800.

\bibitem[{Richemond et~al.(2020)Richemond, Grill, Altché, Tallec, Strub,
  Brock, Smith, De, Pascanu, Piot, and Valko}]{richemond2020byol}
Pierre~H. Richemond, Jean-Bastien Grill, Florent Altché, Corentin Tallec,
  Florian Strub, Andrew Brock, Samuel Smith, Soham De, Razvan Pascanu, Bilal
  Piot, and Michal Valko. 2020.
\newblock \href {http://arxiv.org/abs/2010.10241} {Byol works even without
  batch statistics}.

\bibitem[{Russakovsky et~al.(2014)Russakovsky, Deng, Su, Krause, Satheesh, Ma,
  Huang, Karpathy, Khosla, Bernstein, Berg, and
  Fei{-}Fei}]{DBLP:journals/corr/RussakovskyDSKSMHKKBBF14}
Olga Russakovsky, Jia Deng, Hao Su, Jonathan Krause, Sanjeev Satheesh, Sean Ma,
  Zhiheng Huang, Andrej Karpathy, Aditya Khosla, Michael~S. Bernstein,
  Alexander~C. Berg, and Li~Fei{-}Fei. 2014.
\newblock \href {http://arxiv.org/abs/1409.0575} {Imagenet large scale visual
  recognition challenge}.
\newblock \emph{CoRR}, abs/1409.0575.

\bibitem[{Sermanet et~al.(2017)Sermanet, Lynch, Hsu, and Levine}]{8014803}
Pierre Sermanet, Corey Lynch, Jasmine Hsu, and Sergey Levine. 2017.
\newblock \href {https://doi.org/10.1109/CVPRW.2017.69} {Time-contrastive
  networks: Self-supervised learning from multi-view observation}.
\newblock In \emph{2017 IEEE Conference on Computer Vision and Pattern
  Recognition Workshops (CVPRW)}, pages 486--487.

\bibitem[{Shah et~al.(2023)Shah, Shah, Lau, de~Melo, and Chellapp}]{10031025}
Ketul Shah, Anshul Shah, Chun~Pong Lau, Celso~M. de~Melo, and Rama Chellapp.
  2023.
\newblock \href {https://doi.org/10.1109/WACV56688.2023.00338} {Multi-view
  action recognition using contrastive learning}.
\newblock In \emph{2023 IEEE/CVF Winter Conference on Applications of Computer
  Vision (WACV)}, pages 3370--3380.

\bibitem[{Simonyan and Zisserman(2015)}]{simonyan2015deep}
Karen Simonyan and Andrew Zisserman. 2015.
\newblock \href {http://arxiv.org/abs/1409.1556} {Very deep convolutional
  networks for large-scale image recognition}.

\bibitem[{Soomro et~al.(2012)Soomro, Zamir, and
  Shah}]{DBLP:journals/corr/abs-1212-0402}
Khurram Soomro, Amir~Roshan Zamir, and Mubarak Shah. 2012.
\newblock \href {http://arxiv.org/abs/1212.0402} {{UCF101:} {A} dataset of 101
  human actions classes from videos in the wild}.
\newblock \emph{CoRR}, abs/1212.0402.

\bibitem[{Srivastava et~al.(2015)Srivastava, Mansimov, and
  Salakhutdinov}]{DBLP:journals/corr/SrivastavaMS15}
Nitish Srivastava, Elman Mansimov, and Ruslan Salakhutdinov. 2015.
\newblock \href {http://arxiv.org/abs/1502.04681} {Unsupervised learning of
  video representations using lstms}.
\newblock \emph{CoRR}, abs/1502.04681.

\bibitem[{Szegedy et~al.(2014)Szegedy, Liu, Jia, Sermanet, Reed, Anguelov,
  Erhan, Vanhoucke, and Rabinovich}]{DBLP:journals/corr/SzegedyLJSRAEVR14}
Christian Szegedy, Wei Liu, Yangqing Jia, Pierre Sermanet, Scott~E. Reed,
  Dragomir Anguelov, Dumitru Erhan, Vincent Vanhoucke, and Andrew Rabinovich.
  2014.
\newblock \href {http://arxiv.org/abs/1409.4842} {Going deeper with
  convolutions}.
\newblock \emph{CoRR}, abs/1409.4842.

\bibitem[{Tan and Le(2019)}]{DBLP:journals/corr/abs-1905-11946}
Mingxing Tan and Quoc~V. Le. 2019.
\newblock \href {http://arxiv.org/abs/1905.11946} {Efficientnet: Rethinking
  model scaling for convolutional neural networks}.
\newblock \emph{CoRR}, abs/1905.11946.

\bibitem[{Tao et~al.(2020{\natexlab{a}})Tao, Wang, and
  Yamasaki}]{Tao2020SelfSupervisedVR}
Li~Tao, Xueting Wang, and T.~Yamasaki. 2020{\natexlab{a}}.
\newblock Self-supervised video representation using pretext-contrastive
  learning.
\newblock \emph{ArXiv}, abs/2010.15464.

\bibitem[{Tao et~al.(2020{\natexlab{b}})Tao, Wang, and
  Yamasaki}]{DBLP:journals/corr/abs-2008-02531}
Li~Tao, Xueting Wang, and Toshihiko Yamasaki. 2020{\natexlab{b}}.
\newblock \href {http://arxiv.org/abs/2008.02531} {Self-supervised video
  representation learning using inter-intra contrastive framework}.
\newblock \emph{CoRR}, abs/2008.02531.

\bibitem[{Tian et~al.(2019)Tian, Krishnan, and
  Isola}]{DBLP:journals/corr/abs-1906-05849}
Yonglong Tian, Dilip Krishnan, and Phillip Isola. 2019.
\newblock \href {http://arxiv.org/abs/1906.05849} {Contrastive multiview
  coding}.
\newblock \emph{CoRR}, abs/1906.05849.

\bibitem[{Tian et~al.(2020)Tian, Sun, Poole, Krishnan, Schmid, and
  Isola}]{DBLP:journals/corr/abs-2005-10243}
Yonglong Tian, Chen Sun, Ben Poole, Dilip Krishnan, Cordelia Schmid, and
  Phillip Isola. 2020.
\newblock \href {http://arxiv.org/abs/2005.10243} {What makes for good views
  for contrastive learning}.
\newblock \emph{CoRR}, abs/2005.10243.

\bibitem[{Tong et~al.(2022)Tong, Song, Wang, and Wang}]{tong2022videomae}
Zhan Tong, Yibing Song, Jue Wang, and Limin Wang. 2022.
\newblock \href {http://arxiv.org/abs/2203.12602} {Videomae: Masked
  autoencoders are data-efficient learners for self-supervised video
  pre-training}.

\bibitem[{Touvron et~al.(2021)Touvron, Cord, Sablayrolles, Synnaeve, and
  J{\'{e}}gou}]{DBLP:journals/corr/abs-2103-17239}
Hugo Touvron, Matthieu Cord, Alexandre Sablayrolles, Gabriel Synnaeve, and
  Herv{\'{e}} J{\'{e}}gou. 2021.
\newblock \href {http://arxiv.org/abs/2103.17239} {Going deeper with image
  transformers}.
\newblock \emph{CoRR}, abs/2103.17239.

\bibitem[{Vaswani et~al.(2017)Vaswani, Shazeer, Parmar, Uszkoreit, Jones,
  Gomez, Kaiser, and Polosukhin}]{DBLP:journals/corr/VaswaniSPUJGKP17}
Ashish Vaswani, Noam Shazeer, Niki Parmar, Jakob Uszkoreit, Llion Jones,
  Aidan~N. Gomez, Lukasz Kaiser, and Illia Polosukhin. 2017.
\newblock \href {http://arxiv.org/abs/1706.03762} {Attention is all you need}.
\newblock \emph{CoRR}, abs/1706.03762.

\bibitem[{Vondrick et~al.(2016)Vondrick, Pirsiavash, and
  Torralba}]{DBLP:journals/corr/VondrickPT16}
Carl Vondrick, Hamed Pirsiavash, and Antonio Torralba. 2016.
\newblock \href {http://arxiv.org/abs/1609.02612} {Generating videos with scene
  dynamics}.
\newblock \emph{CoRR}, abs/1609.02612.

\bibitem[{Vondrick et~al.(2018)Vondrick, Shrivastava, Fathi, Guadarrama, and
  Murphy}]{DBLP:journals/corr/abs-1806-09594}
Carl Vondrick, Abhinav Shrivastava, Alireza Fathi, Sergio Guadarrama, and Kevin
  Murphy. 2018.
\newblock \href {http://arxiv.org/abs/1806.09594} {Tracking emerges by
  colorizing videos}.
\newblock \emph{CoRR}, abs/1806.09594.

\bibitem[{Wang et~al.(2020)Wang, Jiao, and
  Liu}]{DBLP:journals/corr/abs-2008-05861}
Jiangliu Wang, Jianbo Jiao, and Yun{-}Hui Liu. 2020.
\newblock \href {http://arxiv.org/abs/2008.05861} {Self-supervised video
  representation learning by pace prediction}.
\newblock \emph{CoRR}, abs/2008.05861.

\bibitem[{Wu et~al.(2018)Wu, Xiong, Yu, and
  Lin}]{DBLP:journals/corr/abs-1805-01978}
Zhirong Wu, Yuanjun Xiong, Stella~X. Yu, and Dahua Lin. 2018.
\newblock \href {http://arxiv.org/abs/1805.01978} {Unsupervised feature
  learning via non-parametric instance-level discrimination}.
\newblock \emph{CoRR}, abs/1805.01978.

\bibitem[{Xu et~al.(2019)Xu, Xiao, Zhao, Shao, Xie, and Zhuang}]{8953292}
Dejing Xu, Jun Xiao, Zhou Zhao, Jian Shao, Di~Xie, and Yueting Zhuang. 2019.
\newblock \href {https://doi.org/10.1109/CVPR.2019.01058} {Self-supervised
  spatiotemporal learning via video clip order prediction}.
\newblock In \emph{2019 IEEE/CVF Conference on Computer Vision and Pattern
  Recognition (CVPR)}, pages 10326--10335.

\bibitem[{Yao et~al.(2020)Yao, Liu, Luo, Zhou, and
  Ye}]{DBLP:journals/corr/abs-2006-11476}
Yuan Yao, Chang Liu, Dezhao Luo, Yu~Zhou, and Qixiang Ye. 2020.
\newblock \href {http://arxiv.org/abs/2006.11476} {Video playback rate
  perception for self-supervisedspatio-temporal representation learning}.
\newblock \emph{CoRR}, abs/2006.11476.

\bibitem[{Yuan et~al.(2021)Yuan, Chen, Wang, Yu, Shi, Tay, Feng, and
  Yan}]{DBLP:journals/corr/abs-2101-11986}
Li~Yuan, Yunpeng Chen, Tao Wang, Weihao Yu, Yujun Shi, Francis E.~H. Tay,
  Jiashi Feng, and Shuicheng Yan. 2021.
\newblock \href {http://arxiv.org/abs/2101.11986} {Tokens-to-token vit:
  Training vision transformers from scratch on imagenet}.
\newblock \emph{CoRR}, abs/2101.11986.

\bibitem[{Zhang et~al.(2022)Zhang, Zhou, Bai, Wang, Zhou, Zhang, and
  Zheng}]{ZHANG2022102160}
Pengcheng Zhang, Lei Zhou, Xiao Bai, Chen Wang, Jun Zhou, Liang Zhang, and Jin
  Zheng. 2022.
\newblock \href {https://doi.org/https://doi.org/10.1016/j.displa.2022.102160}
  {Learning multi-view visual correspondences with self-supervision}.
\newblock \emph{Displays}, 72:102160.

\bibitem[{Zhou et~al.(2021)Zhou, Kang, Jin, Yang, Lian, Hou, and
  Feng}]{DBLP:journals/corr/abs-2103-11886}
Daquan Zhou, Bingyi Kang, Xiaojie Jin, Linjie Yang, Xiaochen Lian, Qibin Hou,
  and Jiashi Feng. 2021.
\newblock \href {http://arxiv.org/abs/2103.11886} {Deepvit: Towards deeper
  vision transformer}.
\newblock \emph{CoRR}, abs/2103.11886.

\end{thebibliography}

\end{document}